\def\year{2020}\relax
\documentclass[letterpaper]{article} 
\usepackage{aaai20}  
\usepackage{times}  
\usepackage{helvet} 
\usepackage{courier}  
\usepackage[hyphens]{url}  
\usepackage{graphicx} 
\usepackage{graphicx}  
\usepackage{amsmath}
\usepackage{xfrac}
\usepackage{multirow}
\usepackage{array}
\usepackage{booktabs}

\title{Learning Inverse Depth Regression for Multi-View Stereo \\ with Correlation Cost Volume }
\author{Qingshan Xu and Wenbing Tao
	\thanks{Corresponding author}\\ 
National Key Laboratory of Science and Technology on Multispectral Information Processing\\ 
School of Artifical Intelligence and Automation, Huazhong University of Science and Technology, China\\
\{qingshanxu, wenbingtao\}@hust.edu.cn 
}
 
\begin{document}

\maketitle

\begin{abstract}
Deep learning has shown to be effective for depth inference in multi-view stereo (MVS). However, the scalability and accuracy still remain an open problem in this domain. This can be attributed to the memory-consuming cost volume representation and inappropriate depth inference.  Inspired by the group-wise correlation in stereo matching, we propose an average group-wise correlation similarity measure to construct a lightweight cost volume. This can not only reduce the memory consumption but also reduce the computational burden in the cost volume filtering. Based on our effective cost volume representation, we propose a cascade 3D U-Net module to regularize the cost volume to further boost the performance. Unlike the previous methods that treat multi-view depth inference as a depth regression problem or an inverse depth classification problem, we recast multi-view depth inference as an inverse depth regression task. This allows our network to achieve sub-pixel estimation and be applicable to large-scale scenes. Through extensive experiments on DTU dataset and Tanks and Temples dataset, we show that our proposed network with Correlation cost volume and Inverse DEpth Regression (CIDER\footnote{Code will be available at \url{https://github.com/GhiXu/CIDER}.}), achieves state-of-the-art results, demonstrating its superior performance on scalability and accuracy. 
\end{abstract}

\section{Introduction}
Multi-view stereo (MVS) has attracted great interest in the past few years for its wide applications in autonomous driving, virtual/augmented reality, 3D printing etc. The goal of MVS is to establish the 3D model of a scene from a collection of 2D images with known camera parameters. Recently, this task always follows a two-stage pipeline: depth map estimation and fusion. Of these two stages, depth map estimation plays an important role in the whole pipeline and many MVS methods~\cite{Zheng2014PatchMatch,Galliani2015Massively,Schonberger2016Pixelwise,Xu2019Multi,Huang2018DeepMVS,Yao2018MVSNet,Yao2019RMVSNet} have put effort into accurate depth sensing. 

The core of depth map estimation is to compute the correspondence of each pixel across different images by measuring the similarity between these pixels. Traditional methods~\cite{Galliani2015Massively,Schonberger2016Pixelwise,Xu2019Multi} depend on hand-crafted similarity metrics, e.g., sum of absolute differences (SAD) and normalized cross correlation (NCC), and thus these metrics are sensitive to textureless areas, reflective surfaces and repetitive patterns. To deal with the above challenges, some methods~\cite{Kolmogorov2002Multi,Hosni2013Fast} resort to regularization technologies, such as graph-cuts and cost filtering. However, these engineered regularization methods still struggle in the above challenging areas.    

To overcome the above difficulties, recent works~\cite{Yao2018MVSNet,Yao2019RMVSNet,Huang2018DeepMVS} leverage deep convolutional neural networks (DCNN) to learn multi-view depth inference. Thanks to their powerful cost volume representation and filtering, these works achieve comparable results to the traditional state-of-the-arts. However, because the cost volume representation is proportional to the model resolution, it always requires taking up a lot of memory~\cite{Huang2018DeepMVS,Yao2018MVSNet}. This greatly limits the use of these methods in large-scale and high-resolution scenarios. To alleviate this problem, R-MVSNet~\cite{Yao2019RMVSNet} utilizes the gated recurrent unit (GRU) to sequentially regularize 2D cost maps along the depth direction. Although this dramatically reduces the memory consumption, it cannot incorporate enough context information over the cost volume like the 3D U-Net in MVSNet~\cite{Yao2018MVSNet}. As a result, the performance of the network self degrades and a traditional variational refinement is required. 

Inspired by the group-wise correlation in stereo matching~\cite{Guo2019Group}, we propose an average group-wise correlation similarity measure to construct a lightweight cost volume representation. Specifically, we first extract the deep features for the reference image and source images. With a differential warping module, we compute compact similarity scores between every source image and the reference image using the group-wise correlation. To aggregate an arbitrary number of neighboring image information, we average the similarity scores of all source images to produce a unified cost feature in the cost volume. In this way, our network can greatly reduce the memory consumption and simultaneously fit in an arbitrary number of neighboring images. Based on this effective cost volume representation, we present a cascade 3D U-Net to learn more context information. This further boosts the performance of our network. 

\begin{figure}[t]
	\centering
	\includegraphics[width=0.9\columnwidth]{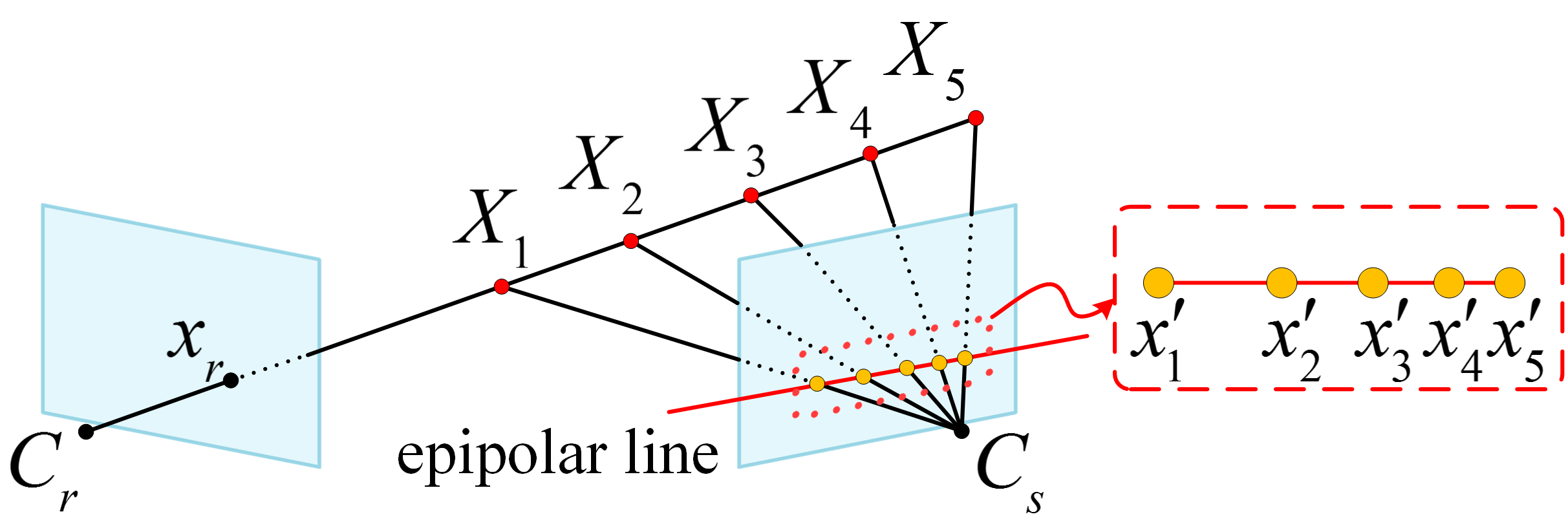}
	\caption{Depth sample and regression. Depth hypotheses are uniformly sampled in depth space. When these uniformly distributed depth hypotheses are projected to a source image, their corresponding 2D points are not distributed uniformly along the epipolar line.}
	\label{depthregression}
\end{figure}

Additionally, existing multi-view depth inference networks~\cite{Huang2018DeepMVS,Yao2018MVSNet,Yao2019RMVSNet} usually cast this task as a depth regression problem or an inverse depth classification problem. It is obvious that the inverse depth classification always introduce stair effects. Thus, DeepMVS~\cite{Huang2018DeepMVS} applies the Fully-Connected Conditional Random Field (DenseCRF)~\cite{Philipp2011Efficient} and R-MVSNet~\cite{Yao2019RMVSNet} utilizes variational refinement to refine their raw predictions. As for the depth regression, this strategy uniformly samples depth hypotheses in depth space and achieves sub-pixel estimation. However, it is not robust. As shown in Figure~\ref{depthregression}, although the sampled depth hypotheses are uniformly distributed in depth space, their projected 2D points in a source image are not distributed uniformly along the epipolar line. Consequently, the true depth hypothesis near the camera center may not be captured by the deep features in a source image. On the other hand, when the true depth hypothesis is far away from the camera center, some depth hypotheses around it will correspond to multiple similar deep features in a source image because these deep features may be sampled from almost the same position. This confuses the true depth hypothesis.

To achieve robust sub-pixel estimation, we cast the multi-view depth inference as an inverse depth regression problem. We sample depth hypotheses in inverse depth space and record their corresponding ordinals. The inverse depth hypotheses are used to obtain the cost map slice in the cost volume while the ordinals are employed to regress the sub-pixel ordinal. The obtained sub-pixel ordinal is further converted to the final depth value, which is utilized to guide the training of our network. The inverse depth regression enables our network to be applied in large-scale scenes.

With the above proposed strategies, our network achieves promising reconstruction results on DTU dataset~\cite{Aanes2016Large} and Tanks and Temples dataset~\cite{Knapitsch2017TTB}. Our contributions are three-fold. 

\begin{itemize}
	\item We propose an average group-wise correlation similarity measure to construct a lightweight cost volume. This greatly eases the memory burden of our network.
	\item We present a cascade 3D U-Net to incorporate more context information to boost the performance of our network.
	\item We treat the multi-view depth inference problem as an inverse depth regression task and demonstrate that the inverse depth regression can reach more robust and accurate results in large-scale scenes. 
\end{itemize}  

\section{Related Work}
Our proposed method is closely related to some learning-based works in stereo matching and multi-view stereo. We briefly review these works in the following.

\textbf{Learning-based Stereo Matching} Stereo matching aims to estimate disparity for a pair of rectified images with small baselines. It can be deemed as a special case of multi-view stereo. With the development of DCNN, many learning-based stereo matching methods have been proposed. {\v{Z}}bontar and LeCun~\cite{Zbontar2015Computing} first introduce a Siamese network to compute matching costs between two image patches. After getting unary features for left and right image patches, these features are concatenated and passed through fully connection layers to predict matching scores. Instead of concatenating unary features, Luo et al.~\cite{Luo2016Efficient} propose a inner product layer to directly correlate unary features. This accelerates the computation of matching cost prediction. In order to achieve end-to-end disparity estimation, DispNet~\cite{Mayer2016Large} is proposed with an encoder-decoder architecture. Kendall et al.~\cite{Kendall2017End} leverage geometry knowledge to form a cost volume by concatenating left and right image features and utilize multi-scale 3D convolutions to regularize the cost volume. Chang and Chen~\cite{Chang2018Pyramid} employ a spatial pyramid  pooling module to incorporate global context information and use a staked hourglass architecture to learn more context information. Tulyakov et al.~\cite{Tulyakov2018Practical} compress the concatenated left-right image descriptors into compact matching signatures to decrease the memory footprint. Guo et al.~\cite{Guo2019Group} propose a group-wise correlation to measure feature similarities, which will not lose too much information like full correlation but reduce the memory consumption and network parameters.

\begin{figure*}[t]
	\centering
	\includegraphics[width=0.9\textwidth]{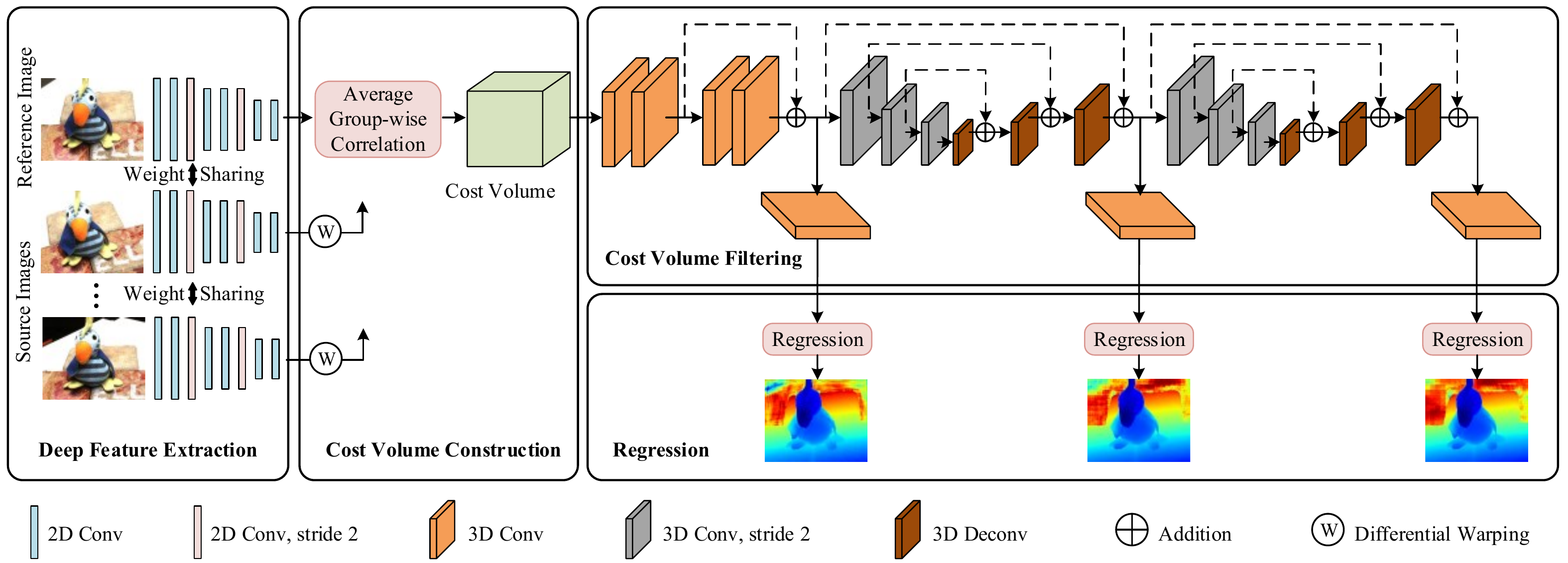}
	\caption{Network architecture of our proposed CIDER. Feature maps are extracted via a weigh-sharing deep feature extraction module for a reference image and source images. The feature maps of source images are warped into the coordinate of the reference image by differential warping. All features are fed to an average group-wise correlation module to construct a lightweight cost volume. The predicted depth maps are obtained by imposing cost volume filtering and regression on the cost volume.}
	\label{network}
\end{figure*}

\textbf{Learning-based Multi-View Stereo} Hartmann et al.~\cite{Hartmann2017Learned} utilize an n-way Siamese network architecture and the mean operation to learn a multi-patch similarity metric. \cite{Ji2017Surfacenet} proposes to encode camera parameters together with images in a 3D voxel representation to learn the 3D model of a scene in an end-to-end manner. Kar et al.~\cite{Kar2017Learning} leverage perspective geometry to construct feature grids for each image and fuse these feature grids into a single grid with a grid recurrent fusion module. Through the 3D grid reasoning, a voxel in the 3D grid can be judged whether it belongs to the surface. DeepMVS~\cite{Huang2018DeepMVS} adopts plane-sweeping to sample image patches and constructs a cost volume for one source image. Then an intra-volume feature aggregation is utilized to perceive non-local information. To tackle an arbitrary number of input images, max-pooling is used to gather the information from neighboring images. DeepMVS poses depth estimation as a multi-label classification problem and thus the DenseCRF is needed to refine the raw predictions. MVSNet~\cite{Yao2018MVSNet} employs a differentiable homography warping to explicitly encode camera geometries in the network and builds the cost volume upon the reference camera frustum. The predictions are regressed from the depth cost volume. To resolve the scalability issue in MVSNet, R-MVSNet~\cite{Yao2019RMVSNet} leverages a gated recurrent unit to sequentially regularize the cost volume. As R-MVSNet also casts the problem as an inverse depth classification task, a variational refinement  is applied to gain more accurate estimation.

\section{Method} 
Our proposed network takes as input a reference image $I_\text{ref}=I_{0}$ and source images $\boldsymbol{I}_\text{src}=\{I_{i}|i=1\dots N-1\}$ with their camera parameters to predict the depth map for the reference image, where $N$ is the total number of input images. As shown in Figure~\ref{network}, our proposed network, CIDER, consists of four modules: deep feature extraction, cost volume construction, cost volume filtering and regression. The detailed parameters of each module are listed in Table~\ref{structure}. 

\subsection{Deep Feature Extraction}
In traditional methods, the original image representations are directly used to construct the cost volume. This may result in the lack of context information in some ambiguous regions, making the depth estimation in these regions failed. Instead, we adopt the multi-scale deep feature extraction network used in MVSNet~\cite{Yao2018MVSNet} to incorporate context information. In this way, for each input image ${3 \times H \times W}$, a multi-scale deep feature ${32 \times \frac{H}{4} \times \frac{W}{4}}$ can be obtained, where $H$ and $W$ are the input image height and width.   

\subsection{Correlation Cost Volume Construction}
After getting the deep features for all input images, we hope to encode these features together with the camera parameters into the network to enable its geometry awareness. Inspired by the traditional plane sweep stereo~\cite{Collins1996Space}, recent learning-based MVS methods, {e.g.}, MVSNet, DeepMVS and R-MVSNet, sample depth hypotheses in 3D space. Based on the sampled depth hypotheses,  the feature representations of source images can be warped into the coordinate of the reference camera to construct a cost volume. Our network also leverages this idea to construct our cost volume. For a pixel $\bf{p}$ in the reference images $I_\text{ref}$, given the $j$-th sampled depth value $d_{j}$ ($j=0\dots D-1$), its corresponding pixel ${\bf p}_{i,j}$ in the source image $I_{i}$ is computed as 
\begin{equation}
{\bf p}_{i,j}={\bf K}_{i}({\bf R }_{\text{ref},i}({\bf K}_\text{ref}^{-1}{\bf p}d_{j})+{\bf t}_{\text{ref},i}),
\end{equation}
where $D$ is the total sample number of depth values, ${\bf K}_{\text{ref}}$ and ${\bf K}_{i}$ are the intrinsic parameters for the reference image $I_\text{ref}$ and the source image $I_{i}$, ${\bf R }_{\text{ref},i}$ is the relative rotation and ${\bf t}_{\text{ref},i}$ is the relative translation. With the above transformation, the deep features of all source images $\boldsymbol{\mathcal{F}}_\text{src}=\{\mathcal{F}_{i}|i=1 \dots N-1\}$ can be warped into the coordinate of the deep feature of the reference image $\mathcal{F}_\text{ref}$. The warped deep features of all source images at depth $d_{j}$ are denoted as $\tilde{\boldsymbol{\mathcal{F}}}_{\text{src},j}=\{\tilde{\mathcal{F}}_{i,j}|i=1 \dots N-1\}$.

In order to measure the multi-view feature similarity, MVSNet~\cite{Yao2018MVSNet} employs a variance-based metric to generate a raw 32-channel cost volume. As the cost volume representation is proportional to the model resolution, it always makes the network have a huge memory footprint. As pointed out in \cite{Yao2018MVSNet}, before feeding the cost volume into the subsequent cost volume regularization module, MVSNet first reduces the 32-channel cost volume to an 8-channel one. Also, the authors of \cite{Tulyakov2018Practical} demonstrate that feeding an 8-channel cost volume which is compressed from a 32-channel cost volume into the regularization module can reach a similar accuracy. This makes us believe that the raw 32-channel cost volume representation may be redundant. Although the above works take the 8-channel cost volume as the input of the cost volume regularization module, they require an extra module to compress the raw 32-channel. This not only increases the computational requirement but also the memory consumption. Thus, we intend to construct a raw 8-channel cost volume to simultaneously reduce the computational requirement and the memory consumption.

Inspired by the group-wise correlation in \cite{Guo2019Group}, we propose an average group-wise correlation similarity measure to construct a lightweight cost volume. Specifically, for the deep reference image feature $\mathcal{F}_\text{ref}$ and the $i$-th warped deep source image feature at depth $d_{j}$,  $\tilde{\mathcal{F}}_{i,j}$, their feature channels are evenly divided into $G$ groups along the channel dimension. Then, the $g$-th group similarity between $\mathcal{F}_\text{ref}$ and $\tilde{\mathcal{F}}_{i,j}$ is computed as 
\begin{equation}
S_{i,j}^{g}=\frac{1}{32/G}\left<\mathcal{F}_\text{ref}^{g}, \tilde{\mathcal{F}}_{i,j}^{g}\right>,
\end{equation}
where $g=0\dots G-1$, $\mathcal{F}_\text{ref}^{g}$ is the $g$-th feature of $\mathcal{F}_\text{ref}$, $\tilde{\mathcal{F}}_{i,j}^{g}$ is the $g$-th feature of $\tilde{\mathcal{F}}_{i,j}$ and $\left<\cdot, \cdot\right>$ is the inner product. When the feature similarities of all $G$ groups are computed for $\mathcal{F}_\text{ref}$ and $\tilde{\mathcal{F}}_{i,j}$, they are packed into a $G$-channel feature similarity map $S_{i,j}$. As there are $D$ sampled depth values, the $D$ feature similarity maps between the reference image and the $i$-th source image are further packed into the cost volume $V_{i}$ of size $G \times \frac{H}{4} \times \frac{W}{4} \times D$. In order to adapt an arbitrary number of input source images, the individual cost volumes for different source images are averaged to compute the following final multi-view cost volume:
\begin{equation}
V=\frac{1}{N-1}\sum_{i=1}^{N-1}V_{i}.
\end{equation} 
Note that, the size of this multi-view cost volume is also $G \times \frac{H}{4} \times \frac{W}{4} \times D$. As aforementioned, a lightweight raw 8-channel cost volume can be obtained by setting $G=8$. This can greatly reduce the memory consumption. Also, this cost volume representation can ease the computation burden of the subsequent cost volume filtering module. 

\begin{table}[!htb]
	\caption{The detailed parameters of the proposed CIDER network. If not specified, each 2D/3D convolution/deconvolution layer is followed by a batch normalization (BN) and a rectified linear unit(ReLU). S1/2 denotes the convolution stride. $*$ denotes no BN and ReLU. $\star$ denotes no ReLU.}
	\centering
	\small
	\begin{tabular}{c|c|c}
		\hline
		Index & Layer Description & Output Size\\
		\hline
		\hline
		& Input Images & H$\times$W$\times$3 \\
		\hline
		\multicolumn{3}{c}{\textbf{Deep Feature Extraction}} \\
		\hline
		1 & Conv2D, 3$\times$3, S1, 8 & H$\times$W$\times$8 \\
		\hline
		2 & Conv2D, 3$\times$3, S1, 8 & H$\times$W$\times$8 \\
		\hline
		3 & Conv2D, 5$\times$5, S2, 16 & \sfrac{1}{2}H$\times$\sfrac{1}{2}W$\times$16 \\
		\hline
		4 & Conv2D, 3$\times$3, S1, 16 & \sfrac{1}{2}H$\times$\sfrac{1}{2}W$\times$16 \\
		\hline
		5 & Conv2D, 3$\times$3, S1, 16 & \sfrac{1}{2}H$\times$\sfrac{1}{2}W$\times$16 \\
		\hline
		6 & Conv2D, 5$\times$5, S2, 32 & \sfrac{1}{4}H$\times$\sfrac{1}{4}W$\times$32 \\
		\hline
		7 & Conv2D, 3$\times$3, S1, 32 & \sfrac{1}{4}H$\times$\sfrac{1}{4}W$\times$32 \\
		\hline
		8 & $\text{Conv2D}^{*}$, 3$\times$3, S1, 32 & \sfrac{1}{4}H$\times$\sfrac{1}{4}W$\times$32 \\
		\hline
		\multicolumn{3}{c}{\textbf{Correlation Cost Volume Construction}} \\
		\hline
		\multicolumn{2}{c|}{Differential Warping and} & \multirow{2}{*}{D$\times$\sfrac{1}{4}H$\times$\sfrac{1}{4}W$\times$8} \\
		\multicolumn{2}{c|}{Average Group-wise Correlation} & \\
		\hline
		\multicolumn{3}{c}{\textbf{Cost Volume Filtering}} \\
		\hline
		9 & Conv3D 3$\times$3$\times$3, S1, 8 & D$\times$\sfrac{1}{4}H$\times$\sfrac{1}{4}W$\times$8 \\
		\hline
		10 & Conv3D 3$\times$3$\times$3, S1, 8 & D$\times$\sfrac{1}{4}H$\times$\sfrac{1}{4}W$\times$8 \\
		\hline
		11 & Conv3D 3$\times$3$\times$3, S1, 8 & D$\times$\sfrac{1}{4}H$\times$\sfrac{1}{4}W$\times$8 \\
		\hline
		\multirow{2}{*}{12} & $\text{Conv3D}^{\star}$ 3$\times$3$\times$3, S1, 8 & \multirow{2}{*}{D$\times$\sfrac{1}{4}H$\times$\sfrac{1}{4}W$\times$8} \\
		& Add the output of 10 & \\
		\hline
		13 & Conv3D 3$\times$3$\times$3, S2, 16 & \sfrac{1}{2}D$\times$\sfrac{1}{8}H$\times$\sfrac{1}{8}W$\times$16 \\ 
		\hline
		14 & Conv3D 3$\times$3$\times$3, S2, 32 & \sfrac{1}{4}D$\times$\sfrac{1}{16}H$\times$\sfrac{1}{16}W$\times$32 \\ 
		\hline
		15 & Conv3D 3$\times$3$\times$3, S2, 64 & \sfrac{1}{8}D$\times$\sfrac{1}{32}H$\times$\sfrac{1}{32}W$\times$64 \\ 
		\hline
		\multirow{2}{*}{16} & Deconv3D 3$\times$3$\times$3, S2, 32 & \multirow{2}{*}{\sfrac{1}{4}D$\times$\sfrac{1}{16}H$\times$\sfrac{1}{16}W$\times$32} \\ 
		& Add the output of 14 & \\
		\hline
		\multirow{2}{*}{17} & Deconv3D 3$\times$3$\times$3, S2, 16 & \multirow{2}{*}{\sfrac{1}{2}D$\times$\sfrac{1}{8}H$\times$\sfrac{1}{8}W$\times$16} \\ 
		& Add the output of 13 & \\
		\hline
		\multirow{2}{*}{18} & Deconv3D 3$\times$3$\times$3, S2, 8 & \multirow{2}{*}{D$\times$\sfrac{1}{4}H$\times$\sfrac{1}{4}W$\times$8} \\ 
		& Add the output of 12 & \\
		\hline
		19-24 & Repeat 13-18 & D$\times$\sfrac{1}{4}H$\times$\sfrac{1}{4}W$\times$8 \\
		\hline
		\multicolumn{3}{c}{\textbf{Regression}} \\
		\hline
		\multirow{2}{*}{25} & From the output of 12 & \multirow{2}{*}{D$\times$\sfrac{1}{4}H$\times$\sfrac{1}{4}W$\times$1} \\
		& $\text{Conv3D}^{*}$ 3$\times$3$\times$3, S1, 1 & \\
		\hline
		& Regression & \sfrac{1}{4}H$\times$\sfrac{1}{4}W \\
		\hline
		\multirow{2}{*}{26} & From the output of 18 & \multirow{2}{*}{D$\times$\sfrac{1}{4}H$\times$\sfrac{1}{4}W$\times$1} \\
		& $\text{Conv3D}^{*}$ 3$\times$3$\times$3, S1, 1 & \\
		\hline
		& Regression & \sfrac{1}{4}H$\times$\sfrac{1}{4}W \\
		\hline
		\multirow{2}{*}{27} & From the output of 24 & \multirow{2}{*}{D$\times$\sfrac{1}{4}H$\times$\sfrac{1}{4}W$\times$1} \\
		& $\text{Conv3D}^{*}$ 3$\times$3$\times$3, S1, 1 & \\
		\hline
		& Regression & \sfrac{1}{4}H$\times$\sfrac{1}{4}W \\
		\hline
	\end{tabular}
	\label{structure}
\end{table}

\subsection{Cost Volume Filtering}
As pointed out in \cite{Kendall2017End}, in order to regress the final sub-pixel estimation, it is import to keep the probability distribution along the depth dimension at each pixel location uni-model. To this end, many works~\cite{Chang2018Pyramid,Guo2019Group,Zhang2019GANet} repeat the same cost volume regularization module to filter cost volumes. Inspired by this idea, we design a cascade 3D U-Net to regularize the above raw 8-channel cost volume. 

Before the cascade 3D U-Net, we set up a residual module and a regression module to let our network learn a better feature representation as \cite{Guo2019Group} does. Then, to handle the depth estimation in some ambiguous regions,  two 3D U-Nets are cascaded to filter the cost volume. Due to the repeated top-down/bottom-up processing structure, our network can learn more context information. The detailed structure of our cost volume filtering module is shown in Figure~\ref{network} and Table~\ref{structure}. Note that, the previous MVS networks never employ this structure due to their large memory consumption caused by the huge cost volume representation, e.g., MVSNet~\cite{Yao2018MVSNet} and R-MVSNet~\cite{Yao2019RMVSNet}. This makes their incorporated context information limited. Thanks to our lightweight cost volume representation, a progressive cost volume filtering can be conducted in our network.  

\subsection{Inverse Depth Regression}
In order to achieve the sub-pixel estimation, \cite{Kendall2017End} first uses disparity regression to estimate the continuous disparity map in stereo matching. As the images are rectified in stereo matching, the uniform disparity sampling in the disparity space results in a uniformly distributed 1D correspondence search problem. Differently, as the images in the multi-view setup are not rectified, the direct depth sampling in the depth space will not lead to the similar distribution in the epipolar line of neighboring images (Figure~\ref{depthregression}). 

As described in the Correlation Cost Volume Construction, the sampled depth hypotheses will be projected to neighboring images to obtain a series of 2D points. To make these 2D points that lie in the same epipolar line distribute as uniformly as possible, discrete depth hypotheses are uniformly sampled in inverse depth space as follows,
\begin{equation}
d_{j}=((\frac{1}{d_{min}}-\frac{1}{d_{max}})\frac{j}{D-1}+\frac{1}{d_{max}})^{-1},j=0\dots D-1,
\end{equation}
where $d_{min}$ and $d_{max}$ are the minimal depth value and the maximal depth value of the reference image. With the above depth value sampling scheme, we can construct a more discriminative cost volume to be sent to the subsequent cost volume filtering module.

As shown in Figure~\ref{network}, there are three output branches in our network. In each branch, the filtered cost volume is converted to a 1-channel cost volume via a 3D convolution operation. To obtain the final continuous depth map, we first regress the sub-pixel ordinal $k$ from the cost volume as follows,
\begin{equation}
k=\sum_{j=0}^{D-1}j\times{p_{j}},
\end{equation}
where $p_{j}$ is the probability at depth value $d_{j}$, which is computed from the predicted cost volume via the softmax function. The final predicted depth value for each pixel is computed as 
\begin{equation}
\hat{d}=((\frac{1}{d_{min}}-\frac{1}{d_{max}})\frac{k}{D-1}+\frac{1}{d_{max}})^{-1}.
\end{equation}
To train our network, we use the ground truth depth map as our supervised signal. We denote the ground truth depth map as $\boldsymbol{d}$ and the three predicted depth maps as $\hat{\boldsymbol{d}}_{0}$, $\hat{\boldsymbol{d}}_{1}$ and $\hat{\boldsymbol{d}}_{2}$. Our final loss function is defined as
\begin{equation}
L=\sum_{q=0}^{2}\lambda_{q}l(\boldsymbol{d},\hat{\boldsymbol{d}}_{q}),
\end{equation}
where $\lambda_{q}$ denotes the weight for the $q$-th predicted depth map and $l(\cdot,\cdot)$ is the mean absolute difference.

\section{Experiments}

In this section, we evaluate our proposed network on DTU dataset~\cite{Aanes2016Large} and Tanks and Temples dataset~\cite{Knapitsch2017TTB}. First, we describe the datasets and evaluation metrics followed by implementation details. Then, we perform ablation studies using Tanks and Temple dataset. Last, we show the benchmarking results on the above datasets.   

\subsection{Datasets and Evaluation Metrics}

\noindent\textbf{DTU Dataset~\cite{Aanes2016Large}} This dataset contains more than 100 object-centric scenes. The ground truth point clouds are scanned in the indoor controlled environments. Thus, The viewpoints and lighting conditions are all deliberately designed. The ground truth camera poses and ground truth point clouds are all publicly available. The image resolution is $1600\times1200$.

\noindent\textbf{Tanks and Temples Dataset~\cite{Knapitsch2017TTB}} This dataset provides both indoor and outdoor scenes. The dataset is further divided into Intermediate datasets and Advanced datasets. Compared to the Intermediate datasets, the Advanced datasets contain larger scale and more complex scenes. Their ground truth camera poses and ground truth point clouds are withheld by the evaluation website. Additionally, this dataset also provides training datasets with their ground truth 3D models available.   

\noindent\textbf{Evaluation Metrics} As suggested in different datasets, the accuracy and the completeness of the distance metric are used for DTU dataset while the accuracy and the completeness of the percentage metric for Tanks and Temple dataset. In order to obtain a summary measure for the accuracy and the completeness, the mean value of them is employed for the distance metric and the $F_{1}$ score is utilized for the percentage metric.

\begin{table*}[!htb]
	\caption{Ablation study results of proposed networks on Tanks and Temples training datasets~\cite{Knapitsch2017TTB} using the percentage metric ($\%$). Due to the GPU memory limitation, image resolution is resized to $1536\times832$ for MVSNet. R-MVSNet$\backslash$Ref. means R-MVSNet without variational refinement.}
	\centering
	\begin{tabular}{ccccccccccc}
		\hline
		Model & Barn & Caterp. & Church & Court. & Ignatius & Meeting. & Truck & Mean & GPU Mem. & Time \\
		\hline
		Base & 24.82 & 6.34 & 39.03 & 37.26 & 11.92 & 22.90 & 12.85 & 22.16 & 11.1 GB  & 2.44 s \\
		AGC & 23.76 & 5.98 & 41.20 & 35.45 &  13.98 & 24.76 & 13.57 & 22.67 & \bf{6.5} GB & \bf{1.90} s \\
		AGC-IDR & 54.00 & 48.39 & 37.48 & 35.25 & 64.26 & 27.89 & 61.48 & 46.96 & \bf{6.5} GB & 2.29 s \\
		CIDER & 56.44 & 49.38 & 40.53 & 36.28 & 64.95 & \bf{29.94} & 63.09 & 48.66 & 7.4 GB & 3.11 s \\
		CIDER (D=256) & \bf{56.97} & 52.62 & 39.47 & 37.38 & \bf{67.71} & 28.52 & \bf{64.56} & \bf{49.60} & 9.6 GB & 4.24 s \\
		\hline
		MVSNet & 24.87 & 6.97 & 37.69 & 35.50 & 11.36 & 21.75 & 17.12 & 22.18 & 11.7 GB & 2.59 s \\
		R-MVSNet$\backslash$Ref. & 51.42 & \bf{53.55} & \bf{45.03} & \bf{40.65} & 67.26 & 23.06 & 62.13 & 49.01 & \bf{6.5} GB & 8.57 s \\
		\hline
	\end{tabular}
	\label{alation}
\end{table*}

\begin{figure*}
	\centering	
	\includegraphics[width=0.136\textwidth]{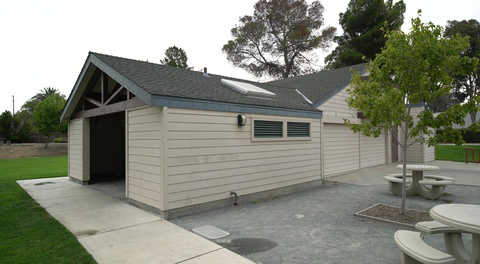} 
	\includegraphics[width=0.136\textwidth]{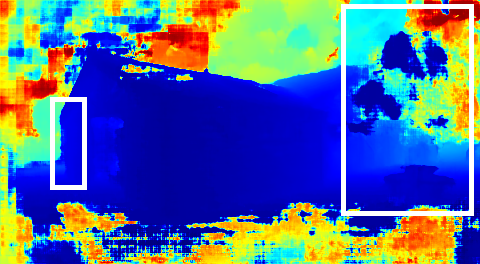}
	\includegraphics[width=0.136\textwidth]{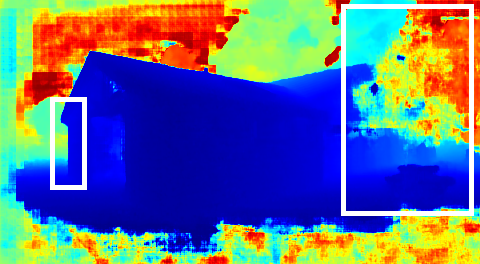}
	\includegraphics[width=0.136\textwidth]{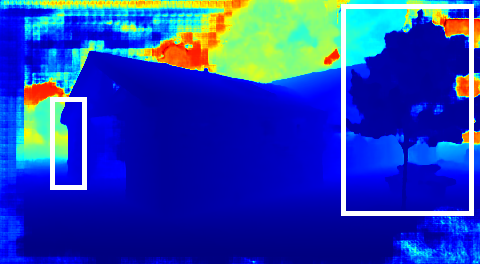}
	\includegraphics[width=0.136\textwidth]{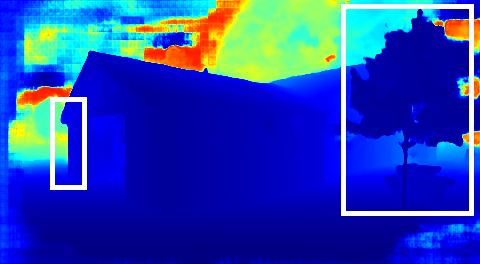}
	\includegraphics[width=0.138\textwidth]{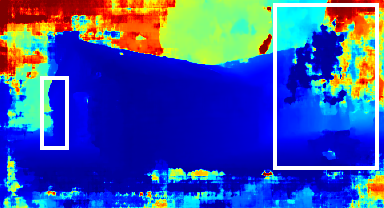}
	\includegraphics[width=0.136\textwidth]{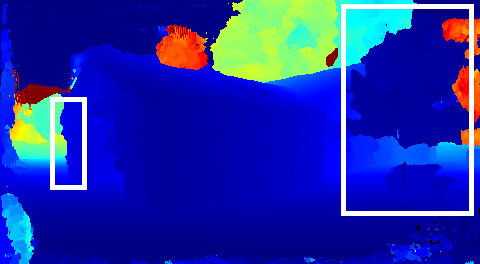}
	\includegraphics[width=0.136\textwidth]{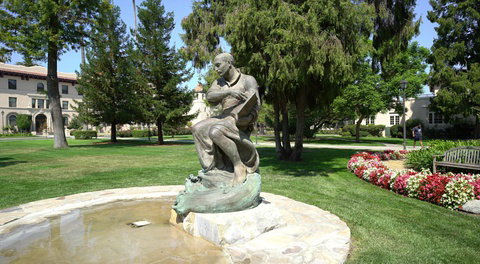} 
	\includegraphics[width=0.136\textwidth]{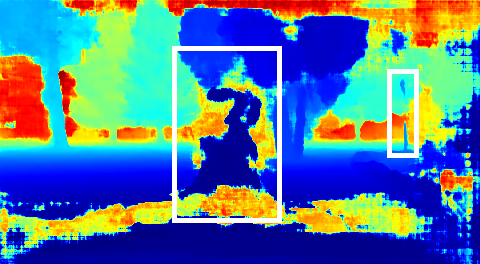}
	\includegraphics[width=0.136\textwidth]{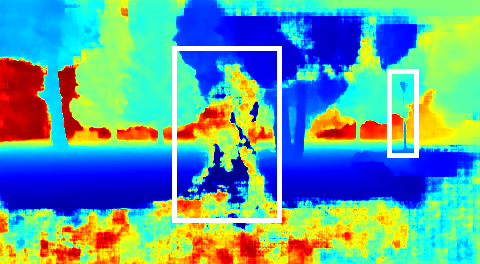}
	\includegraphics[width=0.136\textwidth]{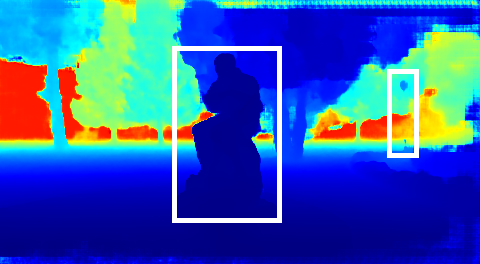}
	\includegraphics[width=0.136\textwidth]{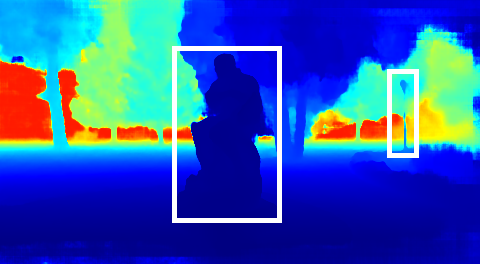}
	\includegraphics[width=0.138\textwidth]{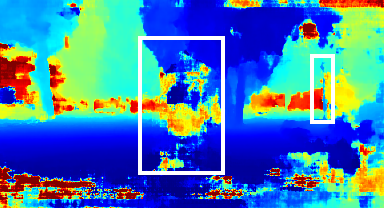}
	\includegraphics[width=0.136\textwidth]{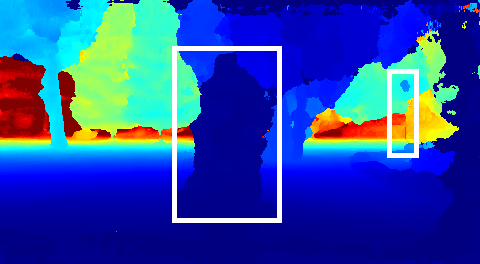}
	\includegraphics[width=0.136\textwidth]{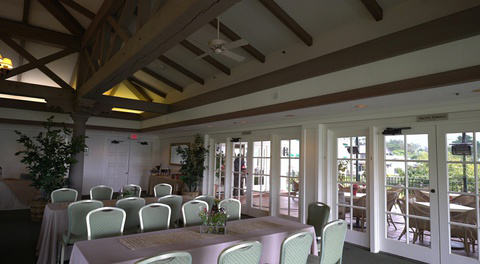} 
	\includegraphics[width=0.136\textwidth]{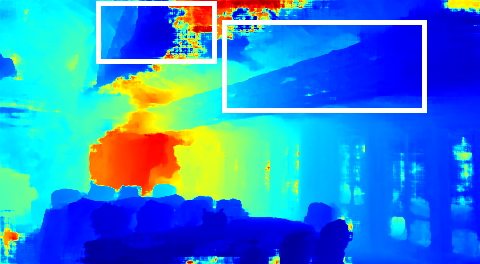}
	\includegraphics[width=0.136\textwidth]{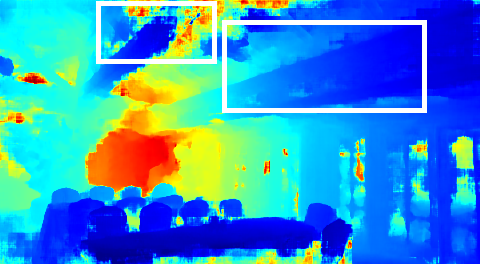}
	\includegraphics[width=0.136\textwidth]{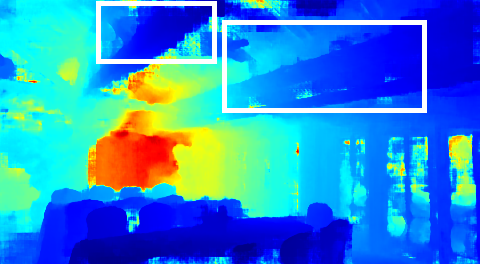}
	\includegraphics[width=0.136\textwidth]{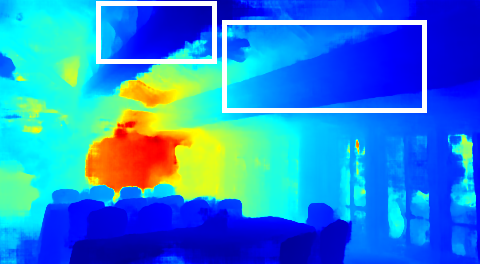}
	\includegraphics[width=0.138\textwidth]{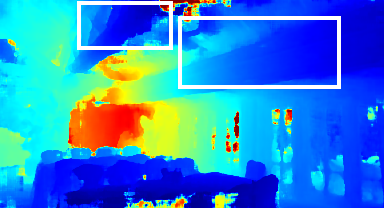}
	\includegraphics[width=0.136\textwidth]{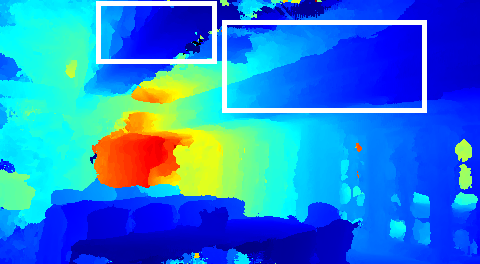}
	\begin{tabular}{p{0.116\textwidth}<{\centering}p{0.116\textwidth}<{\centering}p{0.116\textwidth}<{\centering}p{0.116\textwidth}<{\centering}p{0.118\textwidth}<{\centering}p{0.118\textwidth}<{\centering}p{0.116\textwidth}<{\centering}}
		Image & Base & AGC & AGC-IDR & CIDER & MVSNet & R-MVSNet \\
	\end{tabular}
	\caption{Depth map reconstructions of Barn, Ignatius and Meetingroom, Tanks and Temples training datasets~\cite{Knapitsch2017TTB} using different settings of the proposed networks.}
	\label{AblationDepth}
\end{figure*}

\subsection{Implementation Details}

\noindent\textbf{Training} Following \cite{Ji2017Surfacenet}, we divide the DTU dataset into training set, validation set and test set. We train our network on DTU training set. As DTU dataset does not provide ground truth depth maps, we follow the idea in \cite{Yao2018MVSNet} to generate the depth maps at a resolution of $160\times128$ by leveraging the screened Poisson surface reconstruction~\cite{Kazhdan2013SPS}. During the training, the image size is scaled and cropped to $640\times512$ and the total number of input image is set to $N=3$. $d_{min}$ and $d_{max}$ are fixed to $425mm$ and $935mm$ respectively. The total sample number of depth values is set to $D=192$. The weights for three outputs are set to $\lambda_{0}=0.5$, $\lambda_{1}=0.5$ and $\lambda_{2}=0.7$. We implement our network by using PyTorch. The network is trained for $10$ epoch in total on a TITAN X GPU. We use RMSprop as the optimizer and the initial learning rate is set to $0.001$. The learning rate is decayed every $10,000$ iterations with a base of $0.9$. 

\noindent\textbf{Filtering and Fusion} In order to generate the final single 3D point cloud, we filter and fuse depth maps like other depth map based MVS methods~\cite{Galliani2015Massively,Schonberger2016Pixelwise,Yao2018MVSNet,Xu2019Multi}. Specifically, a probability volume is generated in the regression part of our network. After obtaining the regressed sub-pixel ordinal for each pixel, we locate its corresponding 4-neighboring ordinals and accumulate the probabilities of these ordinals to obtain the final probability representation. This measures the reliability of the depth estimation for each pixel. Then, we filter out the pixels with probability lower than a threshold of $0.8$ to produce a cleaner depth map. In our fusion step, we treat each input image as the reference image in turn. For each pixel in the reference image, we calculate its projected depth and coordinate in neighboring views according to its depth in the reference image. Further, we know the estimated depth in the projected coordinate. Then, we compute the relative depth difference between the projected depth and the estimated depth. With the corresponding depth in neighboring views known, we can compute the reprojected coordinate in the reference image in the same way. We define the distance between the reprojected coordinate and the original coordinate in the reference image as the reprojection error. A pixel will be deemed two-view consistent if its relative depth difference is lower than $0.01$ and its reprojection error is smaller than $1$ pixel. In our experiments, all pixels should be at least three-view consistent and their corresponding 3D points are averaged to produce the final point cloud.

\subsection{Ablation Studies}
In this section, we explore the effectiveness of our proposed strategies, including average group-wise correlation similarity, cascade 3D U-Net filtering and inverse depth regression. To this end, we define a Base model to prove the effectiveness of the above strategies. This Base model replaces the above strategies with variance-based similarity, 3D U-Net filtering and depth regression that are employed in MVSNet~\cite{Yao2018MVSNet}. In order to simultaneously show the generalization of different models on unseen datasets, we use Tanks and Temples training datasets here to conduct experiments. The camera poses are obtained by COLMAP~\cite{Schonberger2016Structure}. The image resolution is resized to $1920\times1056$ as \cite{Yao2018MVSNet,Yao2019RMVSNet} does. The depth sampling number is set to $D=192$ and the input view number is $N=5$\footnote{The discussion of $N$ is shown in supplementary materials.} for all models.

\noindent\textbf{Average Group-wise Correlation Similarity} In order to validate the effectiveness of average group-wise correlation similarity, we replace the variance-based similarity in the Base model with the average group-wise correlation similarity and denote this model as AGC. This makes the cost volume size be reduced from $32\times\frac{H}{4}\times\frac{W}{4}\times{D}$ to $8\times\frac{H}{4}\times\frac{W}{4}\times{D}$. The results are shown in Table~\ref{alation} and Figure~\ref{AblationDepth}. We see that the total memory consumption is reduced by nearly half. Moreover, the reconstruction results of the AGC model remain almost the same as the Base model. This is because our proposed metric also explicitly measures the multi-view feature difference as the variance-based similarity does. As a result, the proposed average group-wise correlation similarity can not only aggregate the multi-view information well but also achieve a compact cost volume representation.  

\noindent\textbf{Inverse Depth Regression} It can be seen from Table~\ref{alation} that the Base model and the AGC model do not generalize well on Tanks and Temples training datasets. As mentioned before, we think that these two models do not carefully consider the epipolar geometry. To prove this, we replace the regression in the AGC model with the inverse depth regression and name this model as AGC-IDR. As shown in Table~\ref{alation}, this model outperforms the previous two models with a significant margin. Moreover, according to the visualization of the reconstructed depth maps in Figure~\ref{AblationDepth}, the network with inverse depth regression can estimate depth maps more accurately than the networks with depth regression. This demonstrates that the inverse depth regression can better depict the distribution of the depth hypotheses in the epipolar line of neighboring images. Therefore, the network can accurately capture the true depth hypothesis.

\noindent\textbf{Cascade 3D U-Net Filtering} In our proposed overall network, CIDER, cascade 3D U-Net filtering is utilized to regularize the cost volume. In Figure~\ref{AblationDepth}, the depth maps reconstructed by AGC-IDR still contain much noise in some ambiguous areas. We suppose that this can be attributed to its limited 3D U-Net regularization and further improve it with cascade 3D U-Net filtering. As shown in Figure~\ref{AblationDepth}, CIDER can better suppress the noise in ambiguous areas than AGC-IDR. Therefore, it achieves better 3D reconstruction results than AGC-IDR, which can be seen from Table~\ref{alation}. It is noteworthy that although two 3D U-Nets are cascaded, the memory consumption is only slightly increased.

In addition, we increase the total depth sampling number from $192$ to $256$ over the same depth range. As illustrated in Table~\ref{alation}, the $F_{1}$ score on Tanks and Temples training datasets is increased from $48.66\%$ to $49.60\%$ and the memory consumption is still acceptable. Thus, we will fix the total depth sampling number to be $256$ when comparing our network with other state of the art learning-based MVS methods on different benchmarks. 

\noindent\textbf{Comparison with Existing Methods} We also compare our method with MVSNet~\cite{Yao2018MVSNet} and R-MVSNet without variational refinement (R-MVSNet$\backslash$Ref.)~\cite{Yao2019RMVSNet}. Table~\ref{alation} shows that our method is much better than MVSNet due to our proposed strategies. Although R-MVSNet$\backslash$Ref. employs the gated recurrent unit to reduce the memory consumption, it cannot incorporate enough context information to tackle the depth estimation in edges and ambiguous regions, e.g., white boxes shown in Figure~\ref{AblationDepth}. Thus, our method is also better than R-MVSNet$\backslash$Ref. As for the running time, due to our proposed lightweight cost volume, the methods with correlation cost volume are faster than R-MVSNet, which acquires larger depth sampling number. 

\begin{table}[t]
	\caption{Quantitative results on the DTU evaluation test~\cite{Aanes2016Large} using the distance metric ($mm$). R-MVSNet$\backslash$Ref. means R-MVSNet without variational refinement.}
	\centering
	\begin{tabular}{cccc}
		\hline
		Method & Mean Acc. & Mean Comp. & Overall \\
		\hline
		SurfaceNet & 0.450 & 1.04 & 0.745 \\
		MVSNet & \bf{0.396} & 0.527 & 0.462 \\
		R-MVSNet$\backslash$Ref. & 0.444 & 0.486 & 0.465 \\
		CIDER & 0.417 & \bf{0.437} & \bf{0.427} \\
		\hline
		R-MVSNet & 0.385 & 0.459 & 0.422 \\
		\hline
	\end{tabular}
	\label{DTUQuant}
\end{table}

\begin{figure}[t]
	\centering	
	\makebox[7pt]{\raisebox{1.8\height}{\rotatebox[origin=c]{90}{Scan 13}}}
	\includegraphics[width=0.47\linewidth]{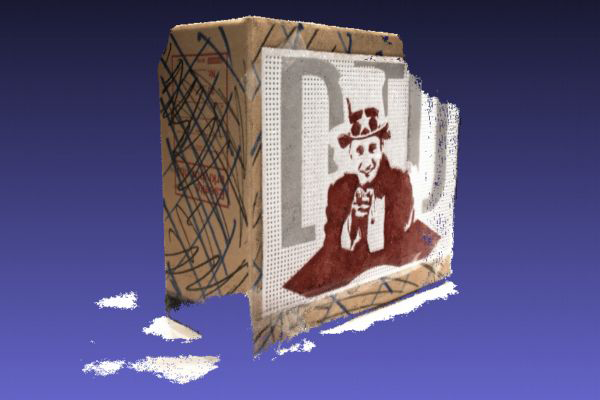} 
	\includegraphics[width=0.47\linewidth]{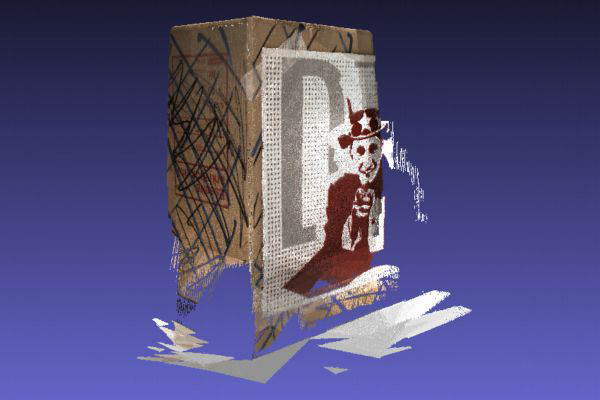}
	\makebox[7pt]{\raisebox{1.8\height}{\rotatebox[origin=c]{90}{Scan 23}}}
	\includegraphics[width=0.47\linewidth]{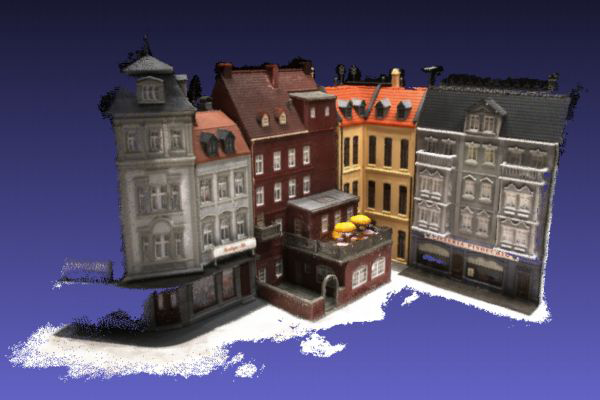} 
	\includegraphics[width=0.47\linewidth]{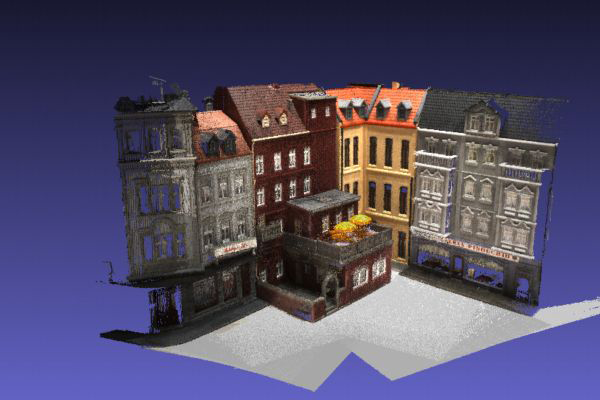}
	\begin{tabular}{p{0.46\linewidth}<{\centering}p{0.46\linewidth}<{\centering}}
		CIDER & Ground Truth \\
	\end{tabular}
	\caption{Our reconstructed point clouds and ground truth on scan 13 and scan 23 of DTU evaluation set~\cite{Aanes2016Large}.}
	\label{DTUQual}
\end{figure}

\begin{table}[t]
	\caption{Quantitative results on the Tanks and Temples Intermediate dataset and Advanced dataset~\cite{Aanes2016Large} using the percentage metric ($\%$).}
	\centering
	\begin{tabular}{c|c|ccc}
		\hline
		Dataset & Method & Acc. & Comp. & $F_{1}$ \\
		\hline
		\multirow{3}{*}{Intermediate} & MVSNet & 40.23 & 49.70 & 43.48 \\
		& R-MVSNet & \bf{43.74} & \bf{57.60} & \bf{48.40} \\
		& CIDER &  42.79 & 55.21 & 46.76 \\
		\hline
		\multirow{3}{*}{Advanced} & MVSNet & - & - & - \\
		& R-MVSNet & \bf{31.47} & \bf{22.05} & \bf{24.91} \\
		& CIDER & 26.64 & 21.27 & 23.12 \\
		\hline
	\end{tabular}
\end{table}

\subsection{Benchmarking}
For the benchmark evaluations, we use our model trained on the DTU training set without fine-tuning. We compare our method with other state of the art learning-based MVS methods, including SurfaceNet~\cite{Ji2017Surfacenet}, MVSNet~\cite{Yao2018MVSNet} and R-MVSNet~\cite{Yao2019RMVSNet}.

\noindent\textbf{DTU Dataset~\cite{Aanes2016Large}} As illustrated in Table~\ref{DTUQuant}, our method produces the best mean completeness and overall score among all methods without post-processing. Figure~\ref{DTUQual} shows the qualitative results of our reconstructions. Note that, the performance of R-MVSNet gains a lot with variational refinement.

\noindent\textbf{Tanks and Temple Dataset~\cite{Knapitsch2017TTB}} In Intermediate dataset, our method surpasses MVSNet by $3.28\%$ and can be applied to large-scale scenes, Advanced datasets while MVSNet cannot. Although R-MVSNet is a little better than our method, we think that its performance advantage comes from its variation refinement post-processing instead of the network self, which can be seen from the evaluation of DTU dataset. See supplementary materials for reconstruction results.

\section{Conclusion}
In this paper, we propose a learning-based multi-view stereo method with correlation cost volume. The correlation cost volume is lightweight due to our proposed average group-wise correlation similarity measure. This reduces the memory consumption and makes our method be scalable on high-resolution images. Moreover, we treat the multi-view depth inference as an inverse depth regression problem. This greatly enhances the generation of our method on unseen large-scale scenarios. We also present a cascade 3D U-Net filtering to improve the accuracy on ambiguous areas. Combined with the above strategies, extensive experiments demonstrate the good applicability of our method, CIDER, on different datasets. 

\noindent\textbf{Acknowledgements} This work was supported by the National Natural Science Foundation of China under Grants 61772213 and 91748204.

\bibliographystyle{aaai}
\bibliography{refs}

\end{document}